\documentclass[journal]{IEEEtran}
\usepackage{cite}

\ifCLASSINFOpdf
\else
\fi
\usepackage{graphicx}
\usepackage{makecell}
\usepackage{xcolor}

\newcommand{\argmin}{\mathop{\mathrm{argmin}}}
\newcommand{\norm}[1]{\Vert#1\Vert}

\begin{document}
\title{Noise Controlled CT Super-Resolution with Conditional Diffusion Model}

\author{Yuang~Wang,
        Siyeop Yoon,
        Rui Hu,
        Baihui Yu,
        Duhgoon Lee,
        Rajiv Gupta,
        Li Zhang,
        Zhiqiang Chen,
        and Dufan~Wu
        % <-this % stops a space
\thanks{Y. Wang, S. Yoon, R. Hu, B. Yu, R. Gupta, and D. Wu are (were) with the Department of Radiology, Massachusetts General Hospital and Harvard Medical School, Boston MA 02114, USA. E-mail: (dwu6@mgh.harvard.edu).}% <-this % stops a space
\thanks{D. Lee is with Neurologica Corp., Danvers MA 01923, USA. }
\thanks{Y. Wang, L. Zhang, and Z. Chen are with the Department of Engineering Physics, Tsinghua University, Beijing 100084, China.}
}

\maketitle
\pagestyle{empty}  % no page number for the second and the later pages
\thispagestyle{empty} % no page number for the first page
\begin{abstract}
Improving the spatial resolution of CT images is a meaningful yet challenging task, often accompanied by the issue of noise amplification. This article introduces an innovative framework for noise-controlled CT super-resolution utilizing the conditional diffusion model. The model is trained on hybrid datasets, combining noise-matched simulation data with segmented details from real data. Experimental results with real CT images validate the effectiveness of our proposed framework, showing its potential for practical applications in CT imaging.
\end{abstract}


% \fi
\IEEEpeerreviewmaketitle

\begin{figure*}
\centering
   \includegraphics[width=18cm]{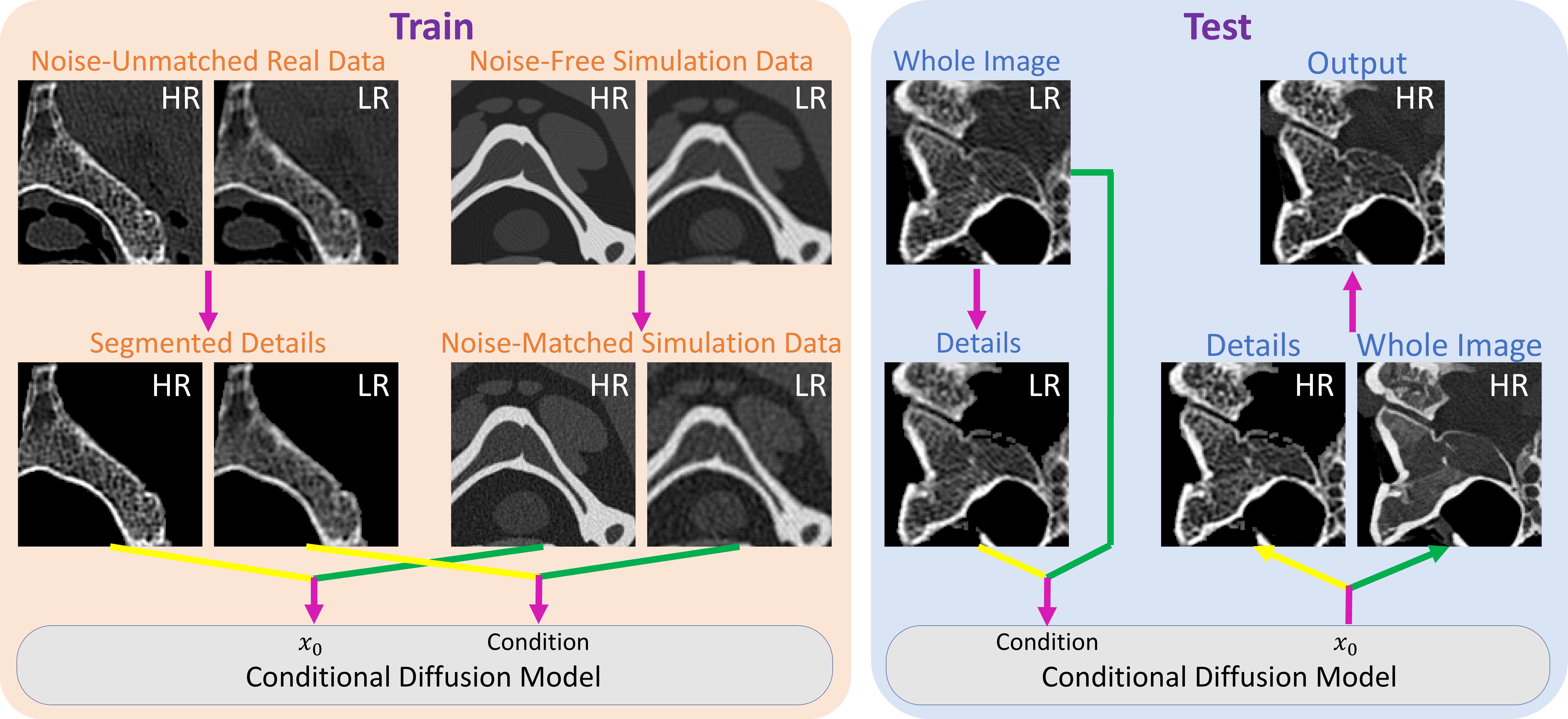}
   \caption
   {Framework of Noise Controlled CT Super-Resolution
   \label{fig:framework}
    }  %note label inside caption
\end{figure*}
\section{Introduction}
\IEEEPARstart{M}{edical} imaging, especially Computed Tomography (CT), is crucial for diagnosing and treating health conditions. Achieving higher resolution in CT scans is an ongoing challenge, and super-resolution techniques play a vital role in enhancing spatial resolution. This improvement promises more detailed information for clinicians, leading to better diagnostic accuracy and improved patient care. Diffusion models \cite{song2019generative, ho2020denoising}, rooted in probabilistic modeling and diffusion processes, offer significant advancements in natural imaging tasks, including super-resolution\cite{choi2021ilvr, saharia2022image, li2022srdiff}. Their stability, unlike Generative Adversarial Networks (GANs), ensures reliable image generation without the pitfalls of mode collapse or unrealistic artifacts. In the context of CT super-resolution, diffusion models stand as a source of inspiration, paving the way for novel developments in medical imaging.

Achieving super-resolution in CT images presents a complex challenge in noise control, particularly when compared to natural images. The training of conditional diffusion models for super-resolution necessitates paired sets of high resolution (HR) and low resolution (LR) CT images. 
Achieving super-resolution without noise amplification requires meticulous matching of noise levels in HR and LR CT images. The scarcity of noise-matched pairs, prompted by radiation exposure, has led certain methodologies to resort to downsampling HR CT images in either the image domain or projection domain to obtain corresponding LR CT images for model training \cite{zhang2021ct}. However, these approaches inadvertently exacerbate noise while improving spatial resolution. Some techniques introduce Gaussian noise to LR CT images\cite{8736838} before or after downsampling to harmonize with the noise level, yet their efficacy in super-resolution may waver when applied to real data due to the introduction of unrealistically distributed noise.

In this article, we present an innovative framework for noise-controlled CT super-resolution, utilizing a conditional diffusion model trained on hybrid datasets. Numerical phantoms are employed to generate noise-matched simulation pairs of HR and LR CT images. Furthermore, details absent in numerical phantoms are segmented from noise-unmatched real pairs and integrated into the training process. Testing using real CT images validates the effectiveness of the proposed framework in real-world scenarios.
\section{Method}

\subsection{Conditional Diffusion Model}
In our article, we employ the Conditional Denoising Diffusion Probabilistic Model (DDPM)\cite{saharia2022image, li2022srdiff} for super-resolution in CT images. Here, the set of LR images serves as the condition $y$, and the set of HR images forms the generation target $x_0$. The Conditional DDPM contains forward and reverse processes. The forward process, a Markov chain, gradually introduces noise into the image until it becomes standard Gaussian noise, with the following process:

\begin{equation}
q\left(x_t|x_{t-1}\right)=N\left(x_t|\sqrt{1-\beta_t}x_{t-1},\beta_tI\right),
\label{eq:q(xt|xt-1)}
\end{equation}
where $t$ is the diffusion time step from 0 to T, and $\beta_t$ is a small positive hyperparameter determining the speed of the diffusion. $x_0$ is the original image and $x_T$ approximately follows the standard normal distribution.

From (\ref{eq:q(xt|xt-1)}), we can derive $q\left(x_t|x_0\right)$ and $q\left(x_{t-1}|x_0,x_t\right)$:

\begin{equation}
q\left(x_t|x_0\right)=N\left(x_t|\sqrt{\gamma_t}x_0,\left(1-\gamma_t\right)I\right),
\label{eq:q(xt|x0)}
\end{equation}
\begin{equation}
q\left(x_{t-1}|x_t,x_0\right)=N\left(x_{t-1}|\widetilde{\mu_t}\left(x_t,x_0\right),\widetilde{\beta_t}I\right),
\label{eq:q(xt-1|xt,x0))}
\end{equation}
where
\begin{equation}
\gamma_t=\prod_{s=1}^{t}\alpha_s,
\label{eq:gamma_t}
\end{equation}
\begin{equation}
\alpha_t = 1-\beta_t,
\label{eq:alpha_t}
\end{equation}
\begin{equation}
\widetilde{\mu_t}\left(x_t,x_0\right)=\frac{\sqrt{\gamma_{t-1}}\beta_t}{1-\gamma_t}x_0+\frac{\sqrt{\alpha_t}\left(1-\gamma_{t-1}\right)}{1-\gamma_t}x_t,
\label{eq:mu_t}
\end{equation}
and
\begin{equation}
\widetilde{\beta_t}=\frac{1-\gamma_{t-1}}{1-\gamma_t}\beta_t.
\label{eq:beta_t}
\end{equation}

The reverse process, a Markov chain aimed at generating $x_0$ from $x_T \sim N(0, I)$, is modeled as a Gaussian process $p_\theta$:
\begin{equation}
p_\theta\left(x_{t-1}|x_t,y\right)=N\left(x_{t-1}|\mu_\theta\left(x_t,y,t\right),\sigma_t^2\right),
\label{eq:p(xt-1|xt,y)}
\end{equation}
where $\sigma_t^2$ is a hyperparameter and is usually set as $\widetilde{\beta_t}$, and $\mu_\theta$ is trained to match $\widetilde{\mu_t}$ to minimize the Kullback–Leibler (K-L) divergence between $p_\theta\left(x_{t-1}|x_t,y\right)$ and $q\left(x_{t-1}|x_t,x_0\right)$:
\begin{equation}
\theta^*=\argmin_\theta E_q\left\{\frac{1}{2\sigma_t^2}\norm{\mu_\theta(x_t,y,t)-\widetilde{\mu_t}(x_t,x_0)}\right\}.
\label{eq:theta_train_by_mu}
\end{equation}

To simplify the training, we can parameterize $\mu_\theta(x_t,y,t)$ as:
\begin{equation}
\mu_\theta(x_t,y,t)=\widetilde{\mu_t}(x_t,\widehat{x}_{0,\theta}(x_t, y, t)),
\label{eq:mu_theta_reparam}
\end{equation}

where
\begin{equation}
\widehat{x}_{0,\theta}\left(x_t,y,t\right)=\frac{1}{\sqrt{\gamma_t}}\left(x_t-\sqrt{1-\gamma_t}\epsilon_\theta\left(x_t,y,t\right)\right).
\label{eq:x_0_hat}
\end{equation}

It leads to our training loss function
\begin{equation}
\theta^*=\argmin_\theta E_{x_0,y}E_{t,\epsilon}\norm{\epsilon_\theta\left(x_t,y,t\right)-\epsilon}_2^2,
\label{eq:training_loss}
\end{equation}
where
\begin{equation}
x_t=\sqrt{\gamma_t}x_0 + \sqrt{1 - \gamma_t}\epsilon, \epsilon \sim N(0, I).
\label{eq:x_t}
\end{equation}

After the network $\epsilon_\theta$ is trained, one can predict $x_0$ from $x_T \sim N(0,I)$ and the condition $y$ following (\ref{eq:p(xt-1|xt,y)}).

\begin{figure*}
\centering
   \includegraphics[width=15.5cm]{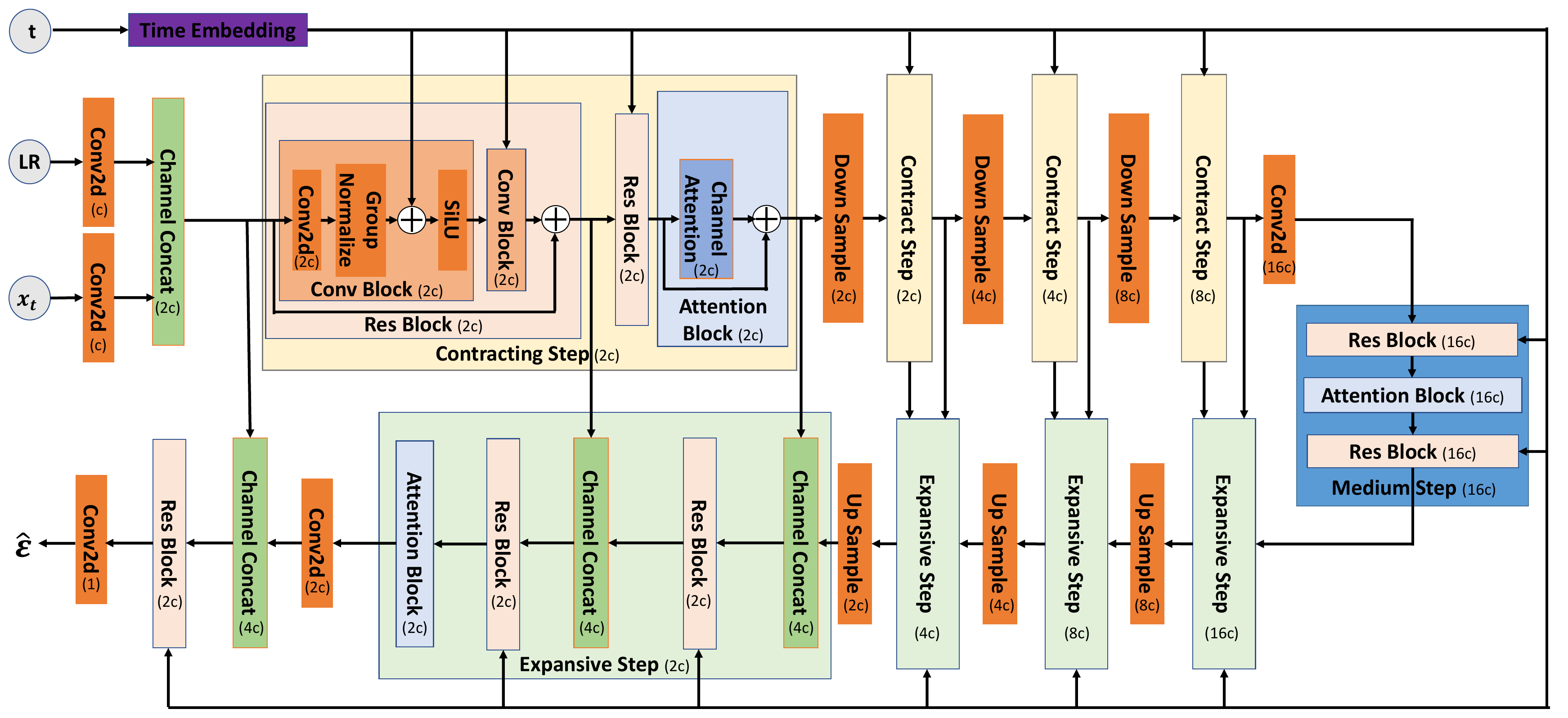}
   \caption
   {Architecture of the conditional noise predictor. The content in parentheses (c, 2c, 4c, 8c and 16c) after the block name indicates the number of output channels of each block, and in our implementation c is set to 32.
   \label{fig:network}
    }  %note label inside caption
\end{figure*}
\subsection{Framework of Noise-Controlled CT Super Resolution}
The framework for noise-controlled CT super-resolution is illustrated in Fig. \ref{fig:framework}. The HR images are the labels $x_0$, and the corresponding LR images are the conditions $y$. Instead of directly training the conditional diffusion model using noise-unmatched real data, which is prone to amplifying noise while enhancing spatial resolution, our approach integrates training with hybrid datasets. These datasets comprise both noise-matched simulation data and segmented bones from real data. 

Numerical phantoms are utilized to simulate CT scans with the same geometry parameters as real CT images, yielding noise-free projections. Assuming that the quantum noise is Gaussian, the following equations are used to inject correlated noises into the HR and LR images:
\begin{equation}
\widehat{p}_{HR}=p_{HR}+\frac{k_{HR}}{\sqrt{N_{HR}}}z
\label{eq:noise_hr}
\end{equation}
\begin{equation}
\widehat{p}_{LR}=p_{LR}+\frac{k_{LR}}{\sqrt{N_{LR}}}downsample\left(z\right),
\label{eq:noise_lr}
\end{equation}
where $z \sim N(0, I)$. $\widehat{p}$ and $p$ are the noisy and noiseless post-log projections, respectively. $N$ is the number of photons per ray that reached the detector, which is different for each detector. $k_{HR}$ and $k_{LR}$ are hyperparameters to match the noise levels of both simulated HR and LR images to that of the real LR image.

Because of the lack of details such as trabecular bone structures in the simulation phantoms, models trained with the simulation data alone would lead to significant oversmooth of the bony structures when applied to the real data. The bone regions from the real data are segmented using threshold followed by hole filling algorithms and opening and closing morphological operations, and serve as part of the training data, as demonstrated in Fig. \ref{fig:framework}. During testing, the trained model is used on both the original LR image and segmented bones. The pixels within the bone mask on the non-segmented super-resolved image are replaced by those in the bone-only super-resolution image. 

In summary, the noise-matched simulated images provide the model the capability of super-resolution without amplification of the noise. Meanwhile, the segmented HR and LR bone pairs encourage the preservation of the detailed bony structures in reality without promoting noise amplification too much. By adopting both training data, the model is trained to enhance spatial resolution without noise amplification and good preservation of the bony structures.

\subsection{Implementation Details}
The real training and testing data were acquired using the OmniTom PCD portable photon counting CT system (Neurologica, Danvers USA) at Massachusetts General Hospital. Acquisition parameters included 120 kVp and 40 mAs for 1$\times$1 binning data, with an effective detector size of 0.12 $\times$ 0.14 mm$^2$ at the isocenter. The projection data were rebinned to 3$\times$3 and 6$\times$5 to generate paired HR and LR real data. The real training dataset comprises 64 slices of a cadaver head, and the real testing dataset includes 32 slices of a separate temporal bone.

Simulation data were generated using the XCIST simulation tool\cite{wu2022xcist}, utilizing head parts from five whole-body NURBS phantoms it provides, resulting in a total of 1040 slices. The hyperparameters $k_{HR}$ and $k_{LR}$ in equations (\ref{eq:noise_hr}) and (\ref{eq:noise_lr}) are set to 0.12 and 0.60, respectively, to match the noise level of the simulation data to that of the real LR image.

Both real and simulation data were reconstructed into 0.3$\times$0.3 mm$^2$ HR images and 0.6$\times$0.6 mm$^2$ LR images using edge-enhancing bone filter. To align their sizes for the conditional diffusion model training, the LR images were upsampled by two times using sinc-interpolation before being fed into the network as condition $y$ \cite{wang2021real}.

The architecture of the conditional noise predictor we employed is illustrated in Fig. \ref{fig:network}. The  core structure of the conditional noise predictor is a U-Net, similar to the one utilized in the DDPM \cite{ho2020denoising}. $y$ and $x_t$ undergo a convolution layer separately and are then concatenated in the channel dimension. This structure is propagated through skip and residual connections across the entire U-Net, allowing the conditional noise predictor to extract information from $y$ and $x_t$ both separately and jointly.

We used $T=1,000$ as the total number of diffusion steps, and $\beta_t$ was selected according to the sigmoid schedule\cite{jabri2022scalable}. The model was trained on randomly cropped patches of 128 $\times$ 128, and tested on the full image. A batch size of 64 was used during the training, with 48 slices from the noise-matched simulation and 16 from the real segmented bone images. The network was trained by the Adam algorithm with a learning rate of $8\times10^{-5}$ for 100,000 iterations.

\section{Result}
\begin{figure*}
\centering
   \includegraphics[width=18cm]{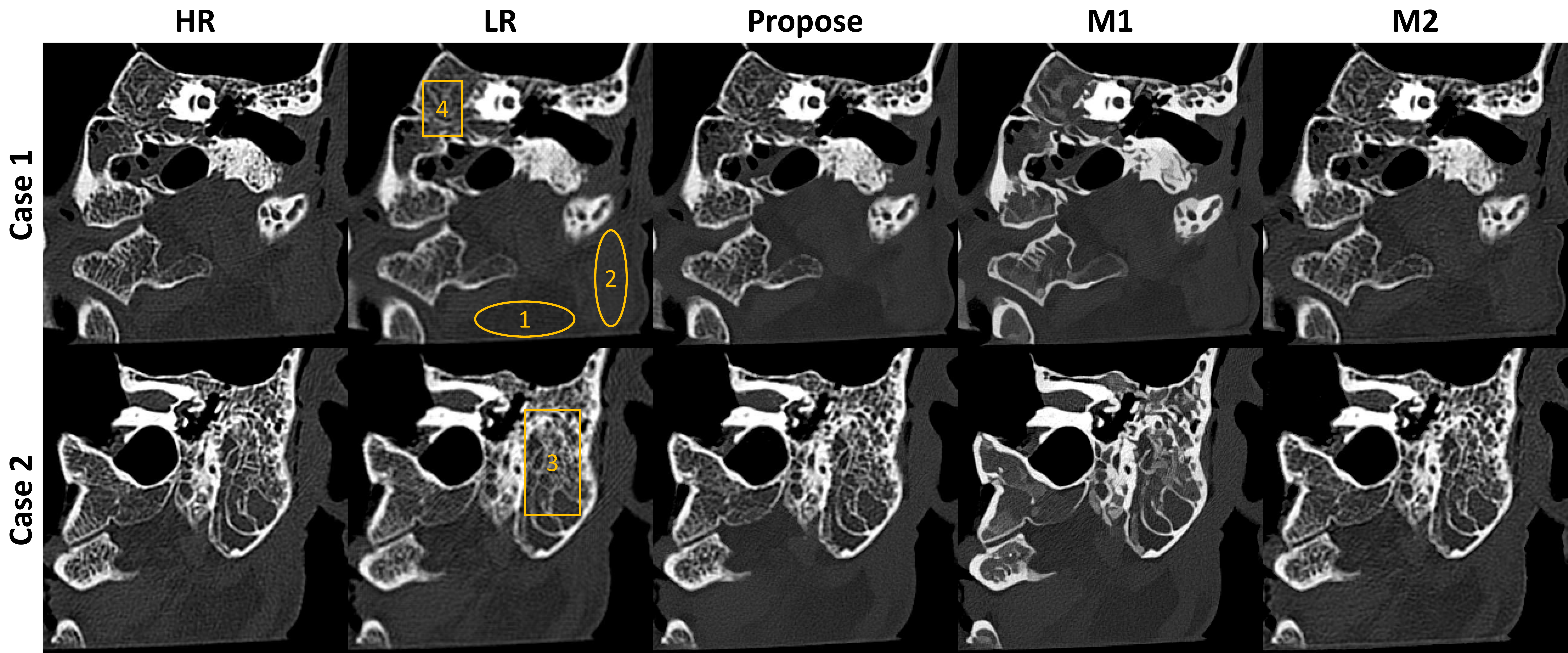}
   \caption
   {Results on LR CT images of a temporal bone using our proposed framework and two comparison methods M1 and M2.
   \label{fig3}
    }  %note label inside caption
\end{figure*}

\begin{table}[htbp]
\begin{center}
\caption{Quantitative analysis for all tested methods on cases and ROIs in Fig. \ref{fig3}. 
\label{table 1}
\vspace*{0.1ex}
}
\begin{tabular} {|l|c|c|c|c|c|c|}
%\begin{edtable}{tabular} {|l|c|c|c|c|}
\hline
Method &\multicolumn{2}{c|}{\makecell[c]{STD (HU)}}&\multicolumn{2}{c|}{\makecell[c]{Haralick feature\\ distance}}&\multicolumn{2}{c|}{PSNR (dB)} \\
\cline{2-7}
        & ROI 1          & ROI 2 & ROI 3          & ROI 4  & Case 1          & Case 2   \\
\hline
        &               &               &               &       & & \\
HR      & 60.4           & 61.2          &  N/A         & N/A   & N/A & N/A        \\
&&&&&&  \vspace{-2mm}\\
LR      & 45.7           & 38.1          & 1367 & 1148    &  23.88         & 24.30     \\
&&&&&&  \vspace{-2mm}\\
Proposed      & 41.5           & 39.9         &325&535     &  23.79         & 24.33     \\
&&&&&&  \vspace{-2mm}\\
M1      & 43.3           & 40.3           &857&976  &  21.46         & 21.62      \\
&&&&&&  \vspace{-2mm}\\
M2      & 54.1          & 51.9          &324&550  &  22.97        & 24.26       \\
        &               &               &               &  &&      \\
\hline
\end{tabular}
%\end{edtable}
\end{center}
\end{table}

The representative results of noise-controlled super resolution from our proposed framework on LR CT images of a temporal bone are presented in Fig. \ref{fig3}, with quantitative analyses in Table \ref{table 1}. The proposed method was further compared to two other methods: M1, which was trained with simulation data only; and M2, which was similar to the proposed method but without matched noise levels in the simulation. 

Regions of interests (ROIs) are selected for quantitative analyses and marked in Fig. \ref{fig3} using yellow ellipses and rectangles. Uniform ROIs 1 and 2 are used to calculate the standard deviation for evaluating the noise level, and ROIs 3 and 4, which contain intricate details, are used to calculate Haralick feature \cite{haralick1973textural} distances to evaluate the similarity of the textures. Peak signal-to-noise ratio (PSNR) is also calculated to assess the consistency of the generated HR images. 

Notably, M1 oversmoothes the trabecular bone structures due to the lack of such structures in the training data. Due to the oversmoothing, it has the worst Haralick feature distances in the bone ROIs (3 and 4) and the worst PSNR compared to the proposed method and M2. M2 has the similar enhancement of the spatial resolution and bony details compared to the proposed method, but it amplifies the noise as shown by the increased standard deviations in ROIs 1 and 2, leading to slightly worse PSNR compared to the proposed method. The proposed method achieved spatial resolution improvement without amplification of the noise. Both M2 and the proposed method had similar PSNR with the LR image, demonstrating that they are not drastically changing the structures as in M1. No significant improvement of the PSNR was observed due to the presence of noise in the HR images.

\section{Conclusion and Discussion}
In conclusion, our proposed framework for noise controlled CT super resolution, leveraging the Conditional DDPM, presents a promising approach to enhance the spatial resolution of CT images while effectively controlling noise. By incorporating segmented details from real data and ensuring a matched noise level in simulation data during training, our model achieved image super-resolution without noise amplification. These results show its potential for practical applications in CT imaging, contributing to advancements in medical imaging technology.

One major limitation of the current method is that the proposed model depends on the precision of details segmentation, and errors in this process may introduce artifacts or result in the loss of details in the final output. Additionally, the training using segmented details featuring a completely black background may pose challenges to the performance of the conditional diffusion model. In our future work, we aim to investigate more robust and effective approaches to seamlessly integrate details unique to real data with noise-matched simulation data, thereby further enhancing the efficacy of noise-controlled CT super-resolution.

% or


%\appendices

%\section{}

%\section*{Acknowledgment}


\ifCLASSOPTIONcaptionsoff
  \newpage
\fi



\bibliographystyle{IEEEtran}
\bibliography{IEEEabrv}
\end{document}